\PassOptionsToPackage{prologue,table,xcdraw}{xcolor}
\PassOptionsToPackage{prologue,table,xcdraw}{color}
\documentclass[sigconf]{acmart}
\usepackage{graphicx}
\usepackage{microtype}
\usepackage{balance}
\usepackage{stfloats}
\usepackage{makecell}
\usepackage{hyperref}
\usepackage{url}
\usepackage{graphicx}
\usepackage{wrapfig}
\usepackage{booktabs}
\usepackage{multirow}
\usepackage{algpseudocode}
\usepackage[ruled,vlined]{algorithm2e}
\usepackage{amsmath}
\usepackage{amsfonts}
\usepackage{multicol}
\usepackage{enumitem}
\usepackage{pifont}
\usepackage{wrapfig}
\usepackage{xcolor}
\usepackage{soul}
\usepackage[most]{tcolorbox}
\usepackage{colortbl}

\definecolor{wrong_action}{HTML}{8faadd}
\definecolor{right_action}{HTML}{f5b183}

\newcommand{\ourmethod}{StepTool}

\AtBeginDocument{%
  }

\copyrightyear{2025}
\acmYear{2025}
\setcopyright{cc}
\setcctype{by}
\acmConference[CIKM '25]{Proceedings of the 34th ACM International Conference on Information and Knowledge Management}{November 10--14, 2025}{Seoul, Republic of Korea}
\acmBooktitle{Proceedings of the 34th ACM International Conference on Information and Knowledge Management (CIKM '25), November 10--14, 2025, Seoul, Republic of Korea}\acmDOI{10.1145/3746252.3761391}
\acmISBN{979-8-4007-2040-6/2025/11}

\begin{document}

\title{StepTool: Enhancing Multi-Step Tool Usage in LLMs via Step-Grained Reinforcement Learning}


\author{Yuanqing Yu}
\affiliation{
  \institution{DCST, Tsinghua University}
  \institution{Quan Cheng Laboratory}
  \city{Beijing 100084}
  \country{China}
  \postcode{100084}
}
\email{yyq23@mails.tsinghua.edu.cn}

\author{Zhefan Wang}
\affiliation{
  \institution{DCST, Tsinghua University}
  \city{Beijing 100084}
  \country{China}
  \postcode{100084}
}
\email{wzf23@mails.tsinghua.edu.cn}

\author{Weizhi Ma}
\authornote{Corresponding author. This work is supported by the Natural Science Foundation of China (Grant No.
U21B2026, 62372260), Wuxi Research Institute of Applied Technologies, Tsinghua University. Weizhi Ma is also sponsored by the Beijing Nova Program.}

\affiliation{%
  \institution{AIR, Tsinghua University}
  \city{Beijing 100084}
  \country{China}
  \postcode{100084}
}
\email{mawz@tsinghua.edu.cn}

\author{Shuai Wang}
\affiliation{%
  \institution{Huawei Noah's Ark Lab}
  \city{Beijing 100084}
  \country{China}
  \postcode{100084}
}
\email{wangshuai231@huawei.com}

\author{Chuhan Wu}
\affiliation{%
  \institution{Huawei Noah's Ark Lab}
  \city{Beijing 100084}
  \country{China}
  \postcode{100084}
}
\email{wuchuhan15@gmail.com}

\author{Zhiqiang Guo}
\affiliation{
  \institution{DCST, Tsinghua University}
  \city{Beijing 100084}
  \country{China}
  \postcode{100084}
}
\email{georgeguo.gzq.cn@gmail.com}

\author{Min Zhang}
\authornotemark[1]
\affiliation{%
  \institution{DCST, Tsinghua University}
  \institution{Quan Cheng Laboratory}
  \city{Beijing 100084}
  \country{China}
  \postcode{100084}
}
\email{z-m@tsinghua.edu.cn}

\renewcommand{\shortauthors}{Yuanqing Yu et al.}

\begin{abstract}
Despite their powerful text generation capabilities, large language models (LLMs) still struggle to effectively utilize external tools to solve complex tasks, a challenge known as tool learning. Existing methods primarily rely on supervised fine-tuning, treating tool learning as a text generation problem while overlooking the decision-making complexities inherent in multi-step contexts. In this work, we propose modeling tool learning as a dynamic decision-making process and introduce \textbf{StepTool}, a novel step-grained reinforcement learning framework that enhances LLMs' capabilities in multi-step tool use. StepTool comprises two key components: \textit{Step-grained Reward Shaping}, which assigns rewards to each tool interaction based on its invocation success and contribution to task completion; and \textit{Step-grained Optimization}, which applies policy gradient methods to optimize the model across multiple decision steps. Extensive experiments across diverse benchmarks show that StepTool consistently outperforms both SFT-based and RL-based baselines in terms of task \textit{Pass Rate} and \textit{Recall} of relevant tools. Furthermore, our analysis suggests that StepTool helps models discover new tool-use strategies rather than merely re-weighting prior knowledge. These results highlight the importance of fine-grained decision modeling in tool learning and establish StepTool as a general and robust solution for enhancing multi-step tool use in LLMs.
Code and data are available at \href{https://github.com/yuyq18/StepTool}{https://github.com/yuyq18/StepTool}.
\end{abstract}



\maketitle

\section{Introduction}

Large language models (LLMs) have demonstrated remarkable reasoning and inference abilities, achieving impressive performance on a wide range of tasks~\citep{brown2020language, zeng2022glm, gpt42023technical}. However, some complex tasks that require real-time information or domain-specific knowledge often exceed the capabilities of LLMs without external support.
In recent years, tool learning~\citep{qin2024toollearningfoundationmodels, patil2023gorilla, qin2023toolllm} has emerged as a promising solution to enable LLMs to utilize external tools~(APIs) effectively. As illustrated in~\autoref{fig: overall}, tool learning allows LLMs to dynamically select, invoke, and interact with tools, obtaining real-time responses to aid task completion. Through multi-step interactions with external tools, LLMs can gather the necessary information and resources to tackle complex and challenging tasks.

\begin{figure*}[t!]
    \centering
    \includegraphics[width=0.95\linewidth]{Figs/intro.png}
    \caption{Tool learning scenario (\textit{left}) and overall comparison between supervised fine-tuning (SFT) and step-grained reinforcement learning (\textit{right}). SFT optimizes in a text-generation view, while the step-grained method utilizes step-level rewards to optimize the decision-making abilities.}
    \label{fig: overall}
    \vspace{-0.2cm}
\end{figure*}

To enhance the tool usage capabilities of LLMs, most existing approaches rely on supervised fine-tuning (SFT)~\citep{qin2023toolllm, patil2023gorilla}, optimizing models based on expert-generated trajectories. These trajectories typically consist of a user query, multiple tool calls, and the corresponding responses, as illustrated in~\autoref{fig: overall}. The model is subsequently fine-tuned on these expert trajectories, optimizing its performance from a text-generation perspective.
However, these approaches treat tool learning as a text-generation problem, neglecting the decision-making dynamics inherent in tool usage. This oversight limits their ability to model the multi-step decision-making process necessary for complex tasks effectively.

To address this, we propose framing tool learning as a sequential decision-making process optimized through step-grained reinforcement learning (RL). In this perspective, each tool invocation is treated as an action that leads to a state transition, and models are trained based on these action-state transitions to improve decision-making capabilities.
While previous works have explored the application of RL to optimize LLMs for aligning with human preferences (RLHF)~\citep{christiano2017deep, ouyang2022training} or for mathematical reasoning tasks~\citep{lightman2023let, wang2023math, shao2024deepseekmath}, these approaches face several challenges when applied to tool learning:
1) Classic RLHF methods~\citep{christiano2017deep, ouyang2022training} are well suitable for ``prompt-response'' data in a single-step manner. Although mathematical reasoning tasks involve multiple steps, these steps are generated by LLMs. In contrast, tool learning involves multiple decision-making steps, each with real-time feedback from external environments.
2) In tool learning, rewards for each step cannot be solely based on correctness, as is common in mathematical reasoning. Instead, rewards for tool usage should consider not only the success of the tool invocation but also the benefits of each decision step.

In light of these challenges, we propose a novel step-grained reinforcement learning framework for tool learning, \textbf{\ourmethod{}}. It models tool learning as a sequential decision-making process and treats each tool interaction as a critical decision point that directly impacts task completion, as shown in~\autoref{fig: overall}.
Specifically, \ourmethod{} consists of two core components: \textit{Step-grained Reward Shaping} and \textit{Step-grained Optimization}.
For \textit{Step-grained Reward Shaping}, we design rewards at each step based on both the accuracy of tool invocation and the contribution to the overall task completion, taking into account characteristics of intermediate actions in this scenario, i.e., well-defined formats and explicit task objectives. These step-grained rewards offer richer signals for tool learning, effectively guiding the model in decision-making.
For \textit{Step-grained Optimization}, we propose a step-grained reinforcement-based optimization method based on the theory of policy gradient~\citep{williams1992simple, sutton1999policy}. This method ensures adaptability to dynamic and multi-step interactions, addressing the limitations of single-step approaches like RLHF.

In summary, this work makes the following contributions:
\begin{itemize}[leftmargin=*, topsep=0pt, parsep=0pt, itemsep=2pt]
\item We propose to formulate tool learning as a sequential decision-making process, where each tool interaction is a critical decision point. This perspective enables learning from action-state transitions and facilitates step-level supervision of decision-making.
\item We introduce \ourmethod{}, a novel step-grained reinforcement learning framework for tool learning, which is comprised of step-grained reward shaping tailored to tool learning scenarios, and a step-grained optimization method based on policy gradients.
\item Comprehensive experiments on multiple benchmarks demonstrate the effectiveness of \ourmethod{}, with consistent gains in task completion and intermediate tool-use quality, enhancing overall multi-step tool usage in LLMs.
\end{itemize}

\section{Related Work}

\subsection{Tool Learning of LLMs}

Recent advancements in tool-augmented LLMs have improved their ability to utilize external tools for complex tasks. Early research~\citep{chen2022program, shen2024hugginggpt, schick2024toolformer} enabled LLMs to interact with diverse external tools like program executors, search engines, and QA systems.
Building on these efforts, subsequent models~\citep{qin2023toolllm, patil2023gorilla} leveraged extensive APIs from platforms like RapidAPI to train LLaMA models~\citep{touvron2023llama} through supervised fine-tuning (SFT). 
To support such training, several works~\citep{tang2023toolalpaca, abdelaziz2024granite, liu2024apigen} focused on constructing diverse and verifiable datasets tailored for tool-augmented SFT. Beyond data and supervised training, some recent studies explored prompt-based strategies to enhance tool use capabilities~\citep{wang2024llmsimaginarium}.
A parallel line of work~\citep{liu2024toolnet, zheng2024toolrerank, liu2024toolplanner} investigated the tool retrieval subtask—i.e., retrieving relevant tools from a large candidate pool. These approaches focus primarily on improving retrieval strategies, rather than addressing the full task-solving pipeline.
More recently, \citet{chen2024advancing} applied Direct Preference Optimization (DPO)~\citep{rafailov2024DPO} to tool learning, but neglects the quality of intermediate decisions.
In contrast, our work explicitly defines step-grained rewards and leverages them for step-grained reinforcement-based optimization. This design enables more precise guidance during training and leads to improved performance on multi-step tool-based tasks.

\subsection{Reinforcement Learning for Multi-Step Textual Tasks}

Recent studies have applied reinforcement learning to align LLM agents with multi-step textual tasks~\citep{carta2023grounding, tan2024true, zhou2024archer, wen2024reinforcing, song2024trial}, including embodied reasoning, game-based dialogue, and instruction following. These methods typically operate in constrained or synthetic environments (as represented by ALFWorld~\citep{ALFWorld20} and TwentyQuestions~\citep{zhou2024archer}), where the action space is limited and training-time actions often overlap significantly with those during evaluation, thereby reducing the difficulty of generalization. In contrast, \ourmethod{} addresses tool learning scenarios, where the action space is considerably broader and more dynamic, due to massive real-world APIs and open-ended user inputs. From a reward modeling perspective, prior methods often rely on explicit intermediate feedback or action-level supervision~\citep{carta2023grounding, tan2024true}, which are typically unavailable in tool-use settings. 
Some recent works~\citep{zhou2024archer, wen2024reinforcing} mitigate this by estimating intra-action influence via auxiliary models. In contrast, \ourmethod{} introduces a step-grained optimization framework that directly computes token-level advantages, enabling fine-grained credit assignment across both intra- and inter-action levels, without requiring external estimators or predefined subgoals.

\subsection{Process Supervision in LLMs}

Process supervision has been extensively explored to enhance long-chain reasoning in LLMs. Previous studies~\citep{lightman2023let, uesato2022solving, ma2023let, shao2024deepseekmath, wang2023math} leverage pre-trained process reward models and optimize reasoning using RLHF~\citep{ouyang2022training}. Recent advancements, such as step-level preferences in mathematical reasoning~\citep{lai2024step}, apply DPO using step-level correctness. 
Our approach differs in two key ways. First, in mathematical reasoning, a “step” is a text segment generated by an LLM, whereas in our work, it is a real-time interaction with external tools and environments. Second, mathematical rewards focus on correctness against ground truth, while tool learning rewards consider tool success and process gains.

\section{Problem Formulation}
\label{sec:problem_formulation}
In this work, we model tool learning in LLMs as a multi-step decision-making problem, formulated as a Markov Decision Process (MDP).
The MDP is formally defined by the tuple $M = (\mathcal S, \mathcal A, \mathcal P, R, \gamma)$.
Here, $\mathcal{S}$ denotes the state space where each state $s_t$ represents the current context including dialogue history and prior tool usage; $\mathcal{A}$ is the action space, where each action $a_t$ involves invoking a tool with arguments or generating a final response; $\mathcal{P}$ describes the transition dynamics $P(s_{t+1} \mid s_t, a_t)$; $R$ is the reward function assigning a scalar reward $r_t = R(s_t, a_t)$ at each step; and $\gamma$ is the discount factor balancing immediate and future rewards.

Unlike standard RL settings, tool learning in LLMs poses unique challenges. The state is embedded in natural language, the action space involves structured outputs (tools and arguments), and the environment often includes non-deterministic external APIs. Despite these complexities, formulating tool learning as an MDP provides a principled foundation for sequential modeling and policy optimization.

In this setting, we represent a trajectory $\tau = \{s_1, a_1, s_2, a_2, \ldots, s_T, a_T\}$ as a sequence of state-action pairs that captures the decision-making process during tool use.
The LLM's behavior is governed by a policy $\pi_\theta$, parameterized by $\theta$, which selects actions based on the current state. The goal is to maximize the expected return $\overline{R_\theta}$ over trajectories induced by $\pi_\theta$. Following standard policy gradient methods~\citep{williams1992simple}, the optimization objective is:

\begin{table}[t]
\caption{Notations in the paper.}
\label{table:notation}
\begin{tabular}{>{\raggedright}m{1.0 cm} m{6.5 cm}<{\raggedright}} 
    \toprule
    \textbf{Symbol} & \multicolumn{1}{l}{\textbf{Description}}
    \\
    \midrule
    $s_t$         & State at time $t$ \\
    $a_t$        & Action taken at time $t$ \\
    $r_t$        & Reward received at time $t$ \\
    $\gamma$ & Discount factor for future rewards \\
    $\tau$  & Trajectory consisting of state-action pairs: $\tau = \{s_1, a_1, s_2, a_2, \ldots, s_T, a_T\}$ \\
    $\pi_\theta$       & Policy parameterized by $\theta$ representing the tool selection strategy of the LLM \\
    $\overline{R_\theta}$ & Expected return (average reward) over all trajectories under policy $\pi_\theta$ \\
    $G_t^n$ & Estimated return over the next $n$ steps starting from time $t$ \\
    $V(s_t)$ & Value function estimating expected return from state $s_t$\\
    $\hat{A}(s_t, a_t)$ & Advantage function indicating the relative value of action $a_t$ at state $s_t$ \\
    \bottomrule
\end{tabular}    
\end{table}

\begin{equation}
\begin{aligned}
&\nabla \overline{R_\theta} = \sum \limits_{\tau}R(\tau)\nabla \pi_\theta(\tau) \\
&= \mathbb{E}_{\tau \sim \pi_\theta(\tau), (s_t, a_t) \sim \tau}\left[R(\tau) \sum \limits_{t=1}^T \nabla \log \pi_\theta(a_t|s_t)\right],
\end{aligned}
\label{equ:gradient_R}
\end{equation}

where $R(\tau)$ denotes the total reward obtained from trajectory $\tau$, and $\pi_\theta(\tau)$ is the probability of generating $\tau$ under policy $\pi_\theta$.

To improve training stability and sample efficiency, $R(\tau)$ is often replaced with an advantage estimate $\hat{A}(s_t, a_t)$, which reflects how much better an action $a_t$ is compared to the expected value of state $s_t$ under the current policy~\citep{williams1992simple, schulman2017proximal}:

\begin{equation}
\begin{aligned}\label{equ:advantage}
&\hat{A}(s_t,a_t) = G_t^n - V(s_t) \\
&= r_t + \gamma r_{t+1} +\ldots +\gamma^{T-t} r_{T}-V(s_t),
\end{aligned}
\end{equation}

where $G_t^n$ denotes the estimated future return over $n$ steps, and $V(s_t)$ is the value function estimating the expected cumulative reward from state $s_t$ under policy $\pi_\theta$.
All notations are summarized in~\autoref{table:notation}.

\begin{figure*}[t]
    \centering
    \includegraphics[width=0.8\linewidth]{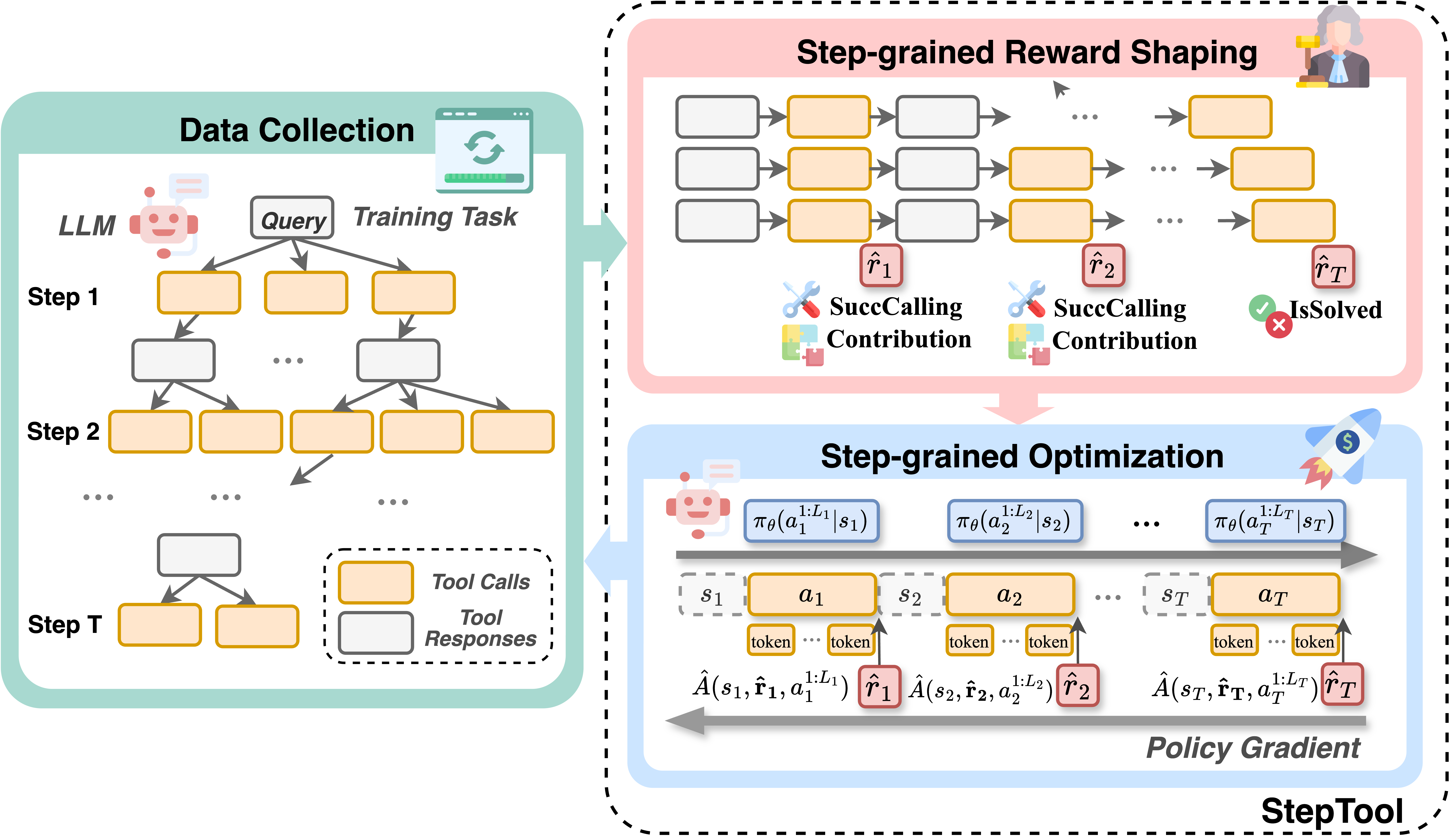}
    \caption{The process flow and architecture of \textbf{\ourmethod{}}. Initially, multi-step tool usage data is collected by performing inference on training tasks with the LLM (\textit{left}).
    With this data, \textit{Step-grained Reward Shaping} is applied to assign rewards at each step of the tool interaction, while \textit{Step-grained Optimization} refines the decision-making process based on policy gradient. (Symbols are described in~\autoref{sec:step-grained optimization} in detail.)}
    \label{fig:our_method}
\end{figure*}

\section{Method}
\ourmethod{} is a step-grained reinforcement learning framework for enhancing multi-step tool use in LLMs. As shown in~\autoref{fig:our_method}, it consists of two components: \textit{Step-grained Reward Shaping} and \textit{Step-grained Optimization}.
We begin by generating tool-use trajectories through inference on training tasks with the backbone LLM. \textit{Step-grained Reward Shaping} assigns feedback to each step based on the correctness of tool invocation and its contribution to task completion. \textit{Step-grained Optimization} then updates the policy using token-level policy gradient methods. In the following sections, we detail the design of each component.

\subsection{\textbf{Step-grained Reward Shaping}}
\label{sec: reward_shaping}
\textit{Step-grained Reward Shaping} provides step-level reward signals for intermediate steps, effectively guiding the model in decision-making.
In tool learning scenarios, where tool invocations follow structured formats and aim to achieve specific task goals, these characteristics naturally support the application of \textit{Step-grained Reward Shaping}, guiding our design approach.

\subsubsection{Step-grained Reward Design}
We design rewards based on two key factors: the success of the tool call (SuccCalling) and its contribution to task completion (Contribution). For the final step, the reward is directly tied to task completion (IsSolved), indicating whether the user's query is resolved.

\paragraph{\textbf{SuccCalling.}} The SuccCalling metric evaluates whether the model successfully executes a tool call with the correct format and content~( i.e., tool name and arguments).
SuccCalling can be formally represented as $\hat{r}^{\mathrm{SC}}_t = \mathrm{SuccCalling}(a_t, s_{t+1})$, where the reward at time $t$ is determined by the action $a_t$ and the subsequent state $s_{t+1}$. 
However, a correct tool call alone does not guarantee task progress. To further guide the model, we introduce the Contribution metric, which assesses how the tool’s action aids the overall task solution.

\paragraph{\textbf{Contribution.}} The Contribution metric evaluates the extent to which the tool’s action facilitates the overall task solution. Actions that contribute minimally, such as redundant steps or irrelevant outputs, receive lower rewards. The $Contribution$ score is based on the relationship between the current action and the final task-solving action, formally defined as $\hat{r}^{\mathrm{Con}}_t = \mathrm{Contribution}(a_t, a_T)$.

\paragraph{\textbf{IsSolved.}} For the final step that gives the final answer, the reward is directly associated with whether the task has been successfully completed. The IsSolved metric evaluates the final answer based on the initial user query, represented as $\hat{r}^{\mathrm{IS}}_t = \mathrm{IsSolved}(q, a_T)$. This reward only depends on the final step and the correctness of the response in addressing the user’s query.

Formally, the reward for each action at step $t$ is defined as:

\begin{equation}
\begin{aligned}
\boldsymbol{\hat{r}_t}=\left\{\begin{array}{l}
\alpha \cdot \hat{r}^{\mathrm{SC}}_t + \hat{r}^{\mathrm{Con}}_t = \alpha \cdot \mathrm{SuccCalling}(a_t, s_{t+1}) \\
+ \mathrm{Contribution}(a_t, a_T), \text{for } t = 1,2,...,T-1 \\
\\
\hat{r}^{\mathrm{IS}}_t = \mathrm{IsSolved}(q, a_T), \text{for } t = T,
\end{array}\right.
\end{aligned}
\label{equ:step_reward}
\end{equation}

where $\alpha$ is a scaling factor balancing the two components.
To ensure consistency, rewards for both intermediate and final steps are normalized to a uniform scale. These rewards can be used to train a reward model or applied directly in reinforcement learning.

\subsubsection{\textbf{Step-grained Reward Acquisition}}
\label{sec: reward_Acq}
Our framework allows for flexible assignment of step-grained rewards through various approaches, including rule-based systems, human annotations, and advanced language models such as GPT-4~\citep{gpt42023technical}. It is important to note that \textbf{the specific reward acquisition method is not the focus of our framework and can be adapted to different settings as needed}.
In our experiments, we adopt a hybrid approach of rule-based heuristics and GPT-4 (\texttt{gpt-4-turbo-2024-04-09}) to annotate step-grained rewards. The detailed process is as follows:
\begin{itemize}[leftmargin=*, topsep=2pt, parsep=2pt]
\item The values of SuccCalling ($r_t^{SC}$) were determined through a combination of rule-based and GPT-4-based annotations. Rules were employed to evaluate the format of the tool invocation (e.g., correctness of tool names and parameters). For invocations meeting the format requirements, GPT-4 was used to ensure the actual success based on the tool response (e.g., a response like \textit{“Since the search parameter is empty, please provide the name of an author…”} indicates a failed invocation due to incorrect arguments).
\item The values of Contribution ($r_t^{Con}$) and IsSolved ($r_t^{IS}$) were obtained directly through automated annotations by GPT-4, which relies on its semantic analysis capabilities.
\end{itemize}

Here, we provide the full GPT-4 annotation prompt used for these rewards in~\autoref{fig:reward_prompt} as a reference.

\begin{tcolorbox}[breakable,title=Annotation Prompt for GPT-4,
colframe=black, coltitle=white, colback=white]
Query: \\
\{query\} \\

Intermediate Steps: \\
\{mid\_steps\} \\
Final Answer: \\
\{final\_answer\} \\

Given the above query, all intermediate steps and the final answer, you need to evaluate the entire task-solving process by following rules: \\
(1) **Successful Tool Calling:** For each intermediate step, determine if a tool was called successfully and give a score of 0 (no) or 1 (yes). \\
(2) **Contribution to Final Answer:** For each intermediate step, rate its contribution to the final answer on a scale from 0 to 5. \\
(3) **Final Answer Status:** Determine if the final answer is ``Solved'',  ``Unsure'', or ``Unsolved''. \\

Now provide your evaluation in JSON format with the parameters of ``succeed\_tool\_calling'', ``contribution\_to\_final\_answer'' and ``final\_answer\_status''  to the function `evaluate\_process\_reward''.\\

\end{tcolorbox}
\begin{figure}[ht]
    \centering
    \vspace{-15pt}
    \caption{
    A Reference Prompt for Step-grained Reward Annotation.
    }
    \label{fig:reward_prompt}
\end{figure}

\subsection{Step-grained Optimization}
\label{sec:step-grained optimization}
Addressing the limitations of the single-step version of RLHF~\citep{ouyang2022training}, we propose a step-grained reinforced optimization strategy based on policy gradient that optimizes all prior steps, ensuring adaptability to multi-step interactions.

\subsubsection{\textbf{Step-grained Optimization Objective}}
Building on the problem formulation (\autoref{sec:problem_formulation}), we now extend the gradient of the expected reward to a token-level consideration. 
Assumed each action $a_t$ consists of a sequence of $L_t$ tokens ($a_t = a_t^{1:L_t}$), the gradient of the expected return $\overline{R_\theta}$ at the step level is expressed as:

\begin{equation}
\begin{aligned}
\nabla \overline{R_\theta} = \mathbb{E}_{\tau \sim \pi_\theta(\tau), (s_t, a_t) \sim \tau}\left[\sum \limits_{t=1}^T \hat{A}(s_t,a_t)  \right.\\
\left. \sum \limits_{i=1}^{L_t} \nabla \log \pi_\theta(a_t^{i}|s_t, a_t^{1:i-1})\right],
\end{aligned}
\label{equ:gradient_R_token}
\end{equation}

where $\hat{A}(s_t,a_t)$ represents the advantage function for the action sequence $a_t$ at step $t$, which is composed of $L_t$ tokens.
Through our \textit{Step-grained Reward Shaping} mechanism, we are able to calculate rewards at each time step $t$ in the trajectory. To better reflect the advantage of each action sequence, we implement the advantage function $\hat{A}(s_t,a_t)$ with our step-grained rewards $\boldsymbol{\hat{r}_t}$ as: 

\begin{equation} 
\begin{aligned}
    \hat{A}(s_t, \boldsymbol{\hat{r}_t},&a_t) = G_t^n - V(s_t) \\
    &= \boldsymbol{\hat{r}_t} + \gamma \boldsymbol{\hat{r}_{t+1}} + \dots + \gamma^{T-t} \boldsymbol{\hat{r}_T} - V(s_t).
\end{aligned}
\label{equ:adv_token} 
\end{equation}

The term $G_t^n$ reflects the cumulative future rewards based on these step-grained rewards $\boldsymbol{\hat{r}_t}$, discounted by factor $\gamma$, extending from step $t$ onward, while $V(s_t)$ is the value function estimating the expected cumulative reward from state $s_t$.

Our optimization objective is thus formalized as:
\begin{equation}
\begin{aligned}
\mathcal L_\theta(\pi)
= \mathbb{E}_{\tau \sim \pi_\theta(\tau), (s_t, a_t) \sim \tau}\left[\sum \limits_{t=1}^T \hat{A}(s_t, \boldsymbol{\hat{r}_t}, a_t) \right.\\
\left. \sum \limits_{i=1}^{L_t}  \log \pi_\theta (a_t^{i}|s_t, a_t^{1:i-1})\right].
\end{aligned}
\label{equ:opt_objective}
\end{equation}

This objective reflects the optimization of the policy $\pi_\theta$ by taking into account the step-level advantage with our step-grained rewards, encouraging the model to select actions that yield higher reward gains.

\paragraph{\textbf{Connection with classic RLHF}}

Additionally, it should be noted that classic RLHF~\citep{ouyang2022training} typically optimizes ``prompt-response'' data with final rewards based on human preferences, which is equivalent to treating the task as a single step ($T=1$).
However, in the scenario of tool learning involving multi-step interactions with external environments, each trajectory consists of multiple intermediate steps. Our method addresses the more complex case of $T > 1$ by applying step-grained rewards and optimizing actions at each step, accounting for both immediate and future outcomes.

\subsubsection{\textbf{An implementation with PPO}}

Notably, our framework is compatible with \textbf{any} policy gradient-based RL algorithm.

As a practical example, we implement the Proximal Policy Optimization (PPO)~\citep{schulman2017proximal} algorithm to demonstrate its application. Here, we estimate the advantage function using Generalized Advantage Estimation (GAE) to improve stability:

\begin{equation}
\begin{aligned}
    \hat{A}(s_t, \boldsymbol{\hat{r}_t}, a_t) &= \delta_t + \cdots + (\gamma \lambda)^{T-t+1}\delta_{T-1}, \\
\text{where } \delta_t &= \boldsymbol{\hat{r}_t} + \gamma V(s_{t+1}) - V(s_t).
\end{aligned}
\end{equation}

To achieve stable training, we employ the PPO-clip version~\citep{schulman2017proximal}, which introduces a clipping mechanism to prevent large updates during optimization. The loss function based on the clipped PPO objective is given by:
\vspace{-0.1cm}
\begin{equation}
\begin{aligned}
\mathcal L^{ppo}_\theta (\pi) = \hat{\mathbb{E}}_{\tau \sim \pi_\theta(\tau), (s_t, a_t) \sim \tau} \left[ \min \left( \sum \limits_{t=1}^T \hat{A}(s_t, \boldsymbol{\hat{r}_t}, a_t) \right.\right.\\
\left.\left. \sum \limits_{i=1}^{L_t} \frac{\log \pi_\theta(a_t^{i}|s_t, a_t^{1:i-1})}{\log \pi_{\theta^{'}}(a_t^{i}|s_t, a_t^{1:i-1})} , \sum \limits_{t=1}^T \hat{A}(s_t, \boldsymbol{\hat{r}_t}, a_t) \right.\right.\\
 \left.\left. \sum \limits_{i=1}^{L_t} \text{clip}\left(\frac{\log\pi_\theta (a_t^{i}|s_t, a_t^{1:i-1})}{\log\pi_{\theta^{'}}(a_t^{i}|s_t, a_t^{1:i-1})}, 1 - \epsilon, 1 + \epsilon \right) \right) \right], 
\end{aligned}
\label{equ:loss_ppo}
\end{equation}

where $\pi_{\theta^{'}}$ represents the old policy used to generate the previous trajectories, and $\epsilon$ is a hyperparameter that controls the allowable deviation between the current and old policies.

To further stabilize training, we introduce a per-token KL divergence penalty between the current and previous policies, as proposed in RLHF~\citep{ouyang2022training}. In our experiments, we apply this PPO-based implementation of StepTool.

\section{Experiments}

In this section, we conduct extensive experiments to answer the following research questions:

\begin{itemize}[leftmargin=*]
    \item \textbf{RQ1:} How does \ourmethod{} perform compared to other optimization methods, including SFT-based and RL-based methods?
    \item \textbf{RQ2:} How does \ourmethod{} perform on different LLM backbone models?
    \item \textbf{RQ3:} Does \ourmethod{} Enable Knowledge Discovery Beyond Prior Re-weighting?
    \item \textbf{RQ4:} How do different components of \ourmethod{} affect?
\end{itemize}

\subsection{Experimental Settings}
\label{sec: experiment}
\subsubsection{\textbf{Benchmark.}}

To assess the multi-step tool usage capabilities of LLMs, we employ two benchmarks that support multi-step tool calls and task completion, StableToolBench~\citep{guo2024stabletoolbench}\footnote{A stable version of ToolBench~\citep{qin2023toolllm}, which includes API response simulations to ensure reliable performance evaluation.} and ToolLens~\citep{qu2024towards-toollens}.
Both benchmarks are designed to test complex task-solving scenarios that require multiple tool calls and intermediate decision-making.
In StableToolBench, each task typically requires 1\textasciitilde6 tool invocations. The benchmark is divided into three subsets (I1., I2., and I3.), each targeting different aspects of tool-use ability. 
For ToolLens, each task involves 1\textasciitilde3 verified tools, and we randomly sample 188 tasks for evaluation. Detailed statistics for both benchmarks are provided in \autoref{tab:task_filteration}.

\begin{table}[h]
\caption{Statistics of test tasks in StableToolBench and ToolLens. \textbf{\# Tasks}, \textbf{\# Rele. Tool} \textbf{\# Rele. Tool / Task} present the number of tasks, the number of relevant tools, and the number of relevant tools for each task, respectively.}
  \centering
  \begin{tabular}{lcccc}
    \toprule
    & \multicolumn{3}{c}{\textbf{ToolBench}} & \textbf{ToolLens} \\
    \cmidrule(lr{0pt}){2-4} \cmidrule(lr{0pt}){5-5}
    & {\textbf{I1.}} & {\textbf{I2.}} & {\textbf{ I3.}} & {\textbf{Test.}} \\
    \midrule
    \textbf{\# Tasks} & 474 & 230 & 61 & 188 \\
    \textbf{\# Rele. Tool} & 1057 & 562 & 180 & 514\\
    \textbf{\# Rele. Tool / Task} & 1\textasciitilde6 & 1\textasciitilde6 & 1\textasciitilde5 & 1\textasciitilde3\\
    \bottomrule
  \end{tabular}
  \label{tab:task_filteration}
\end{table}

\subsubsection{\textbf{Evaluation Metrics.}}
We evaluate performance using two key metrics:
\begin{itemize}[leftmargin=*]
    \item \textbf{Pass (\%)}: the proportion of tasks successfully completed, which reflects the model’s end-to-end task-solving ability.
    \item \textbf{Recall}: the recall of relevant tools used throughout a task, which measures the quality of intermediate decisions for tool selection.
\end{itemize}
Together, these metrics provide a comprehensive view of both the final outcomes and the effectiveness of intermediate tool use.
In addition, we report the \textbf{win rate}~\citep{qin2023toolllm}, an auxiliary metric that measures how often one method outperforms another across individual tasks. This provides further insight into the comparative robustness of different approaches.

\begin{table*}[t]
    \centering
    \caption{Performance comparison between StepTool and other baselines on two benchmarks. We conducted an evaluation three times and reported the average and standard deviation of the results.
    \textbf{Bold} numbers indicate the best performance, and \underline{underlined} numbers denote the second-best. Improvement denotes the improvement of our method over the best result of the baselines. Backbone model is TooLLaMa.
    }   
    \tabcolsep=0.2cm
    \begin{tabular}{llcccccccc}
        \toprule
        \multirow{2}{*}{\textbf{Strategy}} & \multirow{2}{*}{\textbf{Method}} & \multicolumn{2}{c}{\textbf{ToolBench I1.}} & \multicolumn{2}{c}{\textbf{ToolBench I2.}} & \multicolumn{2}{c}{\textbf{ToolBench I3.}} & \multicolumn{2}{c}{\textbf{ToolLens Test.}} \\
        \cmidrule(r){3-4}\cmidrule(r){5-6}\cmidrule(r){7-8}\cmidrule(r){9-10}
         & & Pass (\%) & Recall & Pass (\%) & Recall & Pass (\%) & Recall & Pass (\%) & Recall \\
        \midrule
    \multirow{6}{*}{CoT}
    & SFT~\citep{qin2023toolllm} & 50.6\scriptsize{$\pm{1.6}$} & 0.7952 & 47.1\scriptsize{$\pm{0.8}$} & 0.8081 & \underline{40.4\scriptsize{$\pm{0.8}$}} & 0.6833 & 40.2\scriptsize{$\pm{0.9}$} & 0.6769 \\
    & RFT~\citep{yuan2023scaling} (2023) & 50.2\scriptsize{$\pm{1.2}$} & \underline{0.8061} & 45.9\scriptsize{$\pm{1.8}$} & \underline{0.8197} & 38.5\scriptsize{$\pm{1.2}$} & \underline{0.7536} & 39.5\scriptsize{$\pm{0.7}$} & \underline{0.7323} \\
    & PPO (Final Reward)~\citep{ouyang2022training} & 50.9\scriptsize{$\pm{1.0}$} & 0.8030 & 46.6\scriptsize{$\pm{2.0}$} & 0.8185 & 40.2\scriptsize{$\pm{0.0}$} & 0.6869 & 39.2\scriptsize{$\pm{0.1}$} & 0.6817 \\
    & ETO (DPO)~\citep{song2024trial} (2024) & 50.3\scriptsize{$\pm{0.9}$} &  0.7874 & 45.9\scriptsize{$\pm{0.3}$} & 0.7859 &  38.8\scriptsize{$\pm{1.0}$} & 0.7150  &  40.2\scriptsize{$\pm{0.5}$} & 0.7176 \\
    & ArCHer~\citep{zhou2024archer}  (2024) & \underline{51.8\scriptsize{$\pm{0.8}$}} & 0.8005 & \underline{47.5\scriptsize{$\pm{0.6}$}} & 0.8039 & 35.5\scriptsize{$\pm{2.8}$} & 0.6907 & \underline{42.8\scriptsize{$\pm{0.6}$}} & 0.6693 \\
    \cmidrule{2-10}
    \rowcolor[HTML]{C0F1FF} 
    & \textbf{\ourmethod{} (Ours)}     & \textbf{61.1\scriptsize{$\pm{0.7}$}} & \textbf{0.8743} & \textbf{56.6\scriptsize{$\pm{2.2}$}} & \textbf{0.8992}& \textbf{45.9\scriptsize{$\pm{1.8}$}}& \textbf{0.7724} & \textbf{47.3\scriptsize{$\pm{0.4}$}} & \textbf{0.7500} \\
    \cmidrule{1-10}
    \rowcolor[HTML]{C0F1FF} 
    & \textit{Improvement} & 17.95\% & 8.46\% & 19.16\% & 9.70 \% & 13.61\% & 2.50\% & 10.51\% & 2.42\% \\
    \cmidrule{1-10}
    \multirow{6}{*}{DFSDT}
    & SFT~\citep{qin2023toolllm} & 58.7\scriptsize{$\pm{1.0}$} & 0.8419 & 54.3\scriptsize{$\pm{1.1}$} & 0.8665 & 54.1\scriptsize{$\pm{1.3}$} & 0.7331 & 46.6\scriptsize{$\pm{1.0}$} & 0.7092 \\
    & RFT~\citep{yuan2023scaling} (2023) & 55.0\scriptsize{$\pm{1.6}$} & 0.8490 & 49.5\scriptsize{$\pm{0.8}$} & 0.8531 & \underline{58.5\scriptsize{$\pm{2.4}$}} & \underline{0.7465} & 41.1\scriptsize{$\pm{0.5}$} & \underline{0.7598} \\
    & PPO (Final Reward)~\citep{ouyang2022training} & 59.6\scriptsize{$\pm{1.1}$} & 0.8360 & 54.0\scriptsize{$\pm{1.6}$} & 0.8709 & 39.9\scriptsize{$\pm{0.8}$} & 0.7344 & 42.4\scriptsize{$\pm{0.5}$} & 0.7004 \\
    & ETO (DPO)~\citep{song2024trial} (2024) & 57.1\scriptsize{$\pm{1.2}$} & 0.8412 & 54.5\scriptsize{$\pm{2.1}$} & \underline{0.8747} & 44.0\scriptsize{$\pm{2.8}$} & 0.7298 & \underline{47.9\scriptsize{$\pm{0.8}$}} & 0.7487 \\
    & ArCHer~\citep{zhou2024archer}  (2024) & \underline{60.0\scriptsize{$\pm{1.5}$}} & \underline{0.8491} & \underline{54.5\scriptsize{$\pm{0.8}$}} & 0.8724 & 53.3\scriptsize{$\pm{2.0}$} & 0.7284 & 45.6\scriptsize{$\pm{0.3}$} & 0.7207 \\
    \cmidrule{2-10}
    \rowcolor[HTML]{C0F1FF} 
    & \textbf{\ourmethod{} (Ours)} & \textbf{64.1\scriptsize{$\pm{1.4}$}} & \textbf{0.8797} & \textbf{60.3\scriptsize{$\pm{0.9}$}} & \textbf{0.9004} & \textbf{64.8\scriptsize{$\pm{2.3}$}} & \textbf{0.7831} & \textbf{53.2\scriptsize{$\pm{1.2}$}} & \textbf{0.7819} \\
    \cmidrule{1-10}
    \rowcolor[HTML]{C0F1FF} 
    & \textit{Improvement} & 6.83\% & 3.61 \% & 10.64\% & 2.95 \% & 10.77\% & 4.90 \% & 11.06\% & 2.92 \% \\
    \bottomrule
    \end{tabular}
    \label{tab:main_pass_rate}
\end{table*}

\subsubsection{\textbf{Baselines}} 

To evaluate the effectiveness of our approach, we compare \ourmethod{} with a range of optimization methods for training LLMs, including both SFT-based and RL-based strategies.

For SFT-based optimization methods, we include:
\begin{itemize}[leftmargin=*]
    \item \textbf{SFT}~\citep{qin2023toolllm}, which trains the backbone LLM on expert trajectories using token-level cross-entropy loss.
    \item \textbf{RFT}~\citep{yuan2023scaling} (2023), which augments expert data with self-generated successful trajectories and applies supervised fine-tuning.
\end{itemize}
For RL-based optimization methods, we include:
\begin{itemize}[leftmargin=*]
    \item \textbf{PPO (Final Reward)}~\citep{ouyang2022training}, a single-step RLHF approach that trains LLMs using only the final task reward.
    \item \textbf{ETO (DPO)}~\citep{song2024trial} (2024), which collects both successful and failed trajectories through environment exploration and applies Direct Preference Optimization~\citep{rafailov2024DPO}.
    \item \textbf{ArCHer}~\citep{zhou2024archer}  (2024), a hierarchical multi-turn RL framework designed for training LLM-based agents.
\end{itemize}

We primarily use ToolLLaMa-2-7b-v2 (\textbf{ToolLLaMa})~\citep{qin2023toolllm} as our backbone model, fine-tuned from LLaMA-2-7b-hf and equipped with basic tool-use abilities.
Following the setup in ToolLLaMa, we evaluate all methods under two reasoning strategies: Chain of Thought (CoT)~\citep{wei2022chain} and Depth-First Search Decision Tree (DFSDT)~\citep{qin2023toolllm}.

\subsubsection{\textbf{Training Settings}} 

For SFT-based methods, we use a learning rate of 5e-5, with 4\% warmup and a cosine scheduler, training three epochs. Augmented data for RFT is generated with a temperature of 0.7 and three samples per task.
For ETO (DPO), the DPO loss coefficient is set to $\beta = 0.1$. ArCHer is trained for 100 iterations, with 20 critic epochs per iteration. Both PPO and \ourmethod{} use a learning rate of $1 \times 10^{-5}$, a batch size of 8, and an initial KL coefficient of 0.3.
For a fair comparison, all RL-based methods are trained with the same reward annotation.
In the case of PPO (Final reward) and ArCHer, optimization is carried out using only the final rewards.
For ETO (DPO), additional exploration data is generated at a temperature of 0.7, with three samples collected per task. The final rewards for this exploration data are annotated, and comparative datasets are created accordingly to facilitate DPO optimization.
It is important to distinguish between PPO (Final Reward) and \ourmethod{}. PPO follows the standard RLHF setup~\citep{ouyang2022training}, where rewards are applied only to the final step. In contrast, \ourmethod{} incorporates step-grained rewards, enabling optimization across the entire trajectory.
All experiments are conducted on 4 NVIDIA A100 40G GPUs.
Further experimental details, including hyperparameters, initialization, and training schedules, are available in our code repository\footnote{https://github.com/yuyq18/StepTool} for transparency and reproducibility.

\subsection{RQ1: Overall Performance over other optimization methods}

\autoref{tab:main_pass_rate} presents a comprehensive comparison of \ourmethod{} against strong baselines, evaluated using \textit{Pass (\%)} and \textit{Recall} metrics.
Below are some key observations:

\begin{itemize}[leftmargin=*]
  \item \textbf{Significant improvement in task success.} \ourmethod{} achieves substantial gains in pass rate across all benchmarks, with improvements of up to 19.16\% under the CoT strategy. This demonstrates the effectiveness of our step-wise optimization framework, which is explicitly designed to maximize the overall task-level reward rather than focusing on isolated action steps, thereby better aligning the learning objective with the end-task success.
  \item \textbf{Consistent recall improvement across datasets.} Although recall scores of baseline models are already relatively high, \ourmethod{} achieves consistent improvements across all benchmarks. This suggests that our method enhances not only task success but also the reliability of retrieving relevant tools, reflecting improved alignment with tool-use objectives in multi-step reasoning.
  \item \textbf{Robustness under different reasoning strategies.} \ourmethod{} consistently outperforms all baseline methods under both CoT and DFSDT strategies. Its performance advantage persists even when the reasoning process is enhanced with higher-quality demonstrations, suggesting that our optimization method is robust and complementary to different reasoning strategies.
\end{itemize}

\begin{figure}[t]
    \centering
    \includegraphics[width=1.0\linewidth]{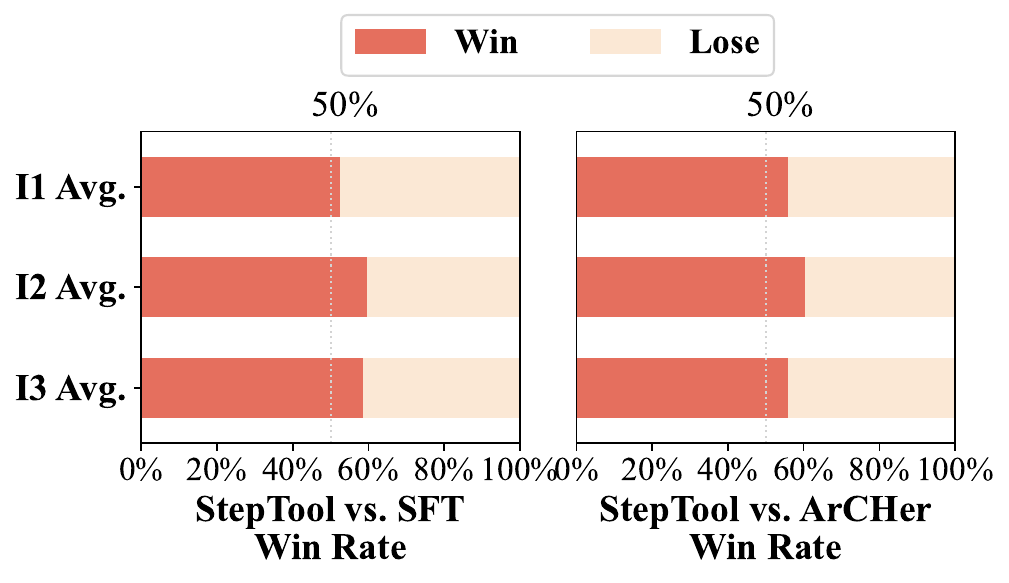}
    \caption{Win rate of \ourmethod{} over leading baselines under the CoT strategy on ToolBench.}
    \label{fig:win_rate}
\end{figure}

In addition, we report win rate~\citep{qin2023toolllm} results under the CoT strategy in~\autoref{fig:win_rate}. Across all evaluation subsets, \ourmethod{} achieves win rates ranging from 52.5\% to 60.3\% when compared to SFT and ArCHer. These results indicate that StepTool consistently generates better solution trajectories, further validating its effectiveness in multi-step, tool-based task solving.

\begin{table}[t]
    \centering
    \caption{The performance on ToolBench I3. and ToolLens Test. on different backbone models.
    ``Llama-3.1-8B'' and ``Qwen2-7B'' stands for \texttt{Llama3.1-8B-Instruct} and \texttt{Qwen2-7B-Instruct} respectively.
    }
    \begin{tabular}{lcccc}
        \toprule
         \multirow{2}{*}{\textbf{Method}} & \multicolumn{2}{c}{\textbf{ToolBench I3.}}  & \multicolumn{2}{c}{\textbf{ToolLens Test.}}\\
        \cmidrule(r){2-3}\cmidrule(r){4-5}
        & Pass (\%) & Recall & Pass (\%) & Recall \\
        \midrule
    \rowcolor[gray]{.8} \multicolumn{5}{c}{\textbf{\textit{Llama-3.1-8B}}}\\
    
    SFT (CoT) & 36.3\scriptsize{$\pm{0.8}$} & 0.6975 & 40.8\scriptsize{$\pm{0.5}$} & 0.6161 \\
    \textbf{Ours (CoT)} & \textbf{42.6\scriptsize{$\pm{2.4}$}} & \textbf{0.7251} & \textbf{51.6\scriptsize{$\pm{0.8}$}} & \textbf{0.6950} \\
    SFT (DFSDT) & 45.9\scriptsize{$\pm{1.3}$} & 0.7271 & 49.6\scriptsize{$\pm{0.5}$} & 0.6857 \\
    \textbf{Ours (DFSDT)} & \textbf{50.5\scriptsize{$\pm{1.0}$}} & \textbf{0.7434} & \textbf{53.3\scriptsize{$\pm{0.3}$}} & \textbf{0.7345}  \\
    \midrule
    \rowcolor[gray]{.8} \multicolumn{5}{c}{\textbf{\textit{Qwen2-7B}}}\\
    SFT (CoT) & 40.7\scriptsize{$\pm{0.8}$} & 0.6866 & 46.6\scriptsize{$\pm{1.2}$} & 0.6037 \\
    \textbf{Ours (CoT)} & \textbf{48.6\scriptsize{$\pm{1.9}$}} & \textbf{0.7230} & \textbf{58.8\scriptsize{$\pm{0.7}$}} & 
\textbf{0.6769} \\
    SFT (DFSDT) & 57.7\scriptsize{$\pm{1.0}$} & 0.7404 & 62.0\scriptsize{$\pm{0.4}$} & 0.7154 \\
    \textbf{Ours (DFSDT)}  & \textbf{62.0\scriptsize{$\pm{1.4}$}} & \textbf{0.7781} & \textbf{63.4\scriptsize{$\pm{1.2}$}} & \textbf{0.7358} \\
         \bottomrule
    \end{tabular}
    \label{tab:base_model}
\end{table}

\subsection{RQ2: Performance on Different Backbone Models}

To further validate the generality and broad applicability of \ourmethod{}, we evaluate its effectiveness on two representative backbone models: Llama3.1-8B-Instruct (\textbf{Llama3.1-8B})\citep{touvron2023llama} and Qwen2-7B-Instruct (\textbf{Qwen2-7B})\citep{yang2024qwen2}. These models are selected due to their widespread adoption in recent open-source LLM research and their strong performance across a variety of downstream tasks.

\autoref{tab:base_model} reports the performance comparison between the standard SFT-trained base models and those optimized with \ourmethod{}. As shown, \ourmethod{} consistently yields improvements in both pass rate and relevant tool recall for Llama3.1-8B and Qwen2-7B across multiple benchmarks.
These consistent gains across different architectures underscore the robustness and versatility of \ourmethod{}. It serves as a model-agnostic enhancement framework, capable of improving multi-step tool usage in LLMs regardless of their architectural design or pretraining origin.

\subsection{RQ3: Does StepTool Enable Knowledge Discovery Beyond Prior Re-weighting?}

To investigate whether \ourmethod{} leads to genuine knowledge discovery or merely re-weights prior knowledge, we evaluate its performance using the Pass@k metric—commonly used in domains such as program synthesis and mathematical reasoning~\citep{ni2022learning, havrilla2024teaching}. A method focused solely on re-weighting priors would show gains predominantly at low values of $k$, with diminishing returns as $k$ increases.
Given the number of sampled trajectories and the cost of real-world API interactions, we randomly selected 120 tasks from ToolBench, uniformly sampling 40 tasks from each of three representative subsets (I1., I2., and I3.) to ensure coverage and diversity.

As shown in~\autoref{tab:pass_k}, StepTool consistently improves performance across all Pass@k metrics. Specifically, it boosts Pass@2 by 13.88\%, Pass@4 by 12.38\%, and Pass@8 by 8.73\% compared to the SFT-trained origin model. These results indicate that the improvements are not confined to just the top few responses, but persist across a broader range of sampled trajectories.
This trend supports the interpretation that StepTool facilitates true knowledge discovery. Rather than merely adjusting prior probabilities, the model benefits from exploration and reward-guided refinement during reinforcement learning, enabling it to uncover new and effective tool-use strategies.

\begin{table}[t]
    \centering
    \caption{Pass@k performance comparison between StepTool and SFT on the ToolLLaMa model, evaluated under the CoT strategy. The results report the average performance across all subsets (I1., I2. and I3.).}
    \renewcommand\tabcolsep{5.5pt}
    \renewcommand\arraystretch{1.0}
    \resizebox{0.45\textwidth}{!}{
    \begin{tabular}{cccr}
    \toprule
    \multirow{2}{*}{\textbf{Metric}} &\multicolumn{2}{c}{\textbf{Methods}} & \multirow{2}{*}{\textbf{Improv.\%}} \\
    \cmidrule{2-3}
         & + SFT & + StepTool &  \\
    \midrule
    Pass Rate@2    & 53.3\scriptsize{$\pm{2.8}$}      & \textbf{60.7\scriptsize{$\pm{2.8}$}}     & 13.88\%  \\
    Pass Rate@4    & 62.2\scriptsize{$\pm{3.8}$}     & \textbf{69.9\scriptsize{$\pm{2.8}$}}    & 12.38\%  \\
    Pass Rate@8    & 67.6\scriptsize{$\pm{3.6}$}     & \textbf{73.5\scriptsize{$\pm{3.3}$}}    & 8.73\%   \\
    \bottomrule
    \end{tabular}
    }
    \label{tab:pass_k}
\end{table}

\begin{figure}[t]
    \centering
    \includegraphics[width=1.0\linewidth]{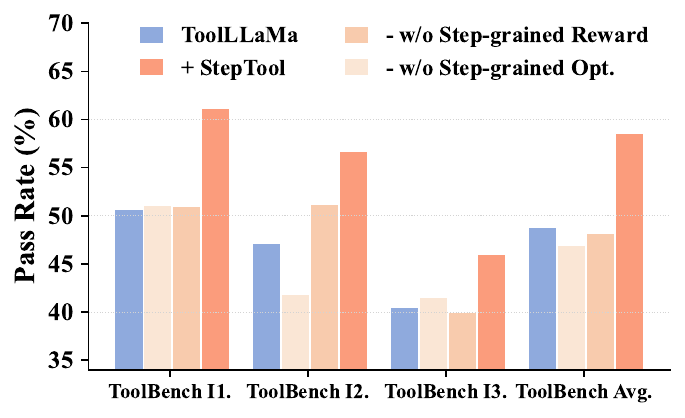}
    \caption{Ablation study on two components of \ourmethod{} under the CoT strategy on ToolBench.
    }
    \label{fig:ablation}
\end{figure}

\subsection{RQ4: Impact of Step-grained Components}

To further evaluate the contributions of each step-grained component in our framework, we conduct an ablation study on two variants of \ourmethod{}.

\textbf{w/o Step-grained Reward.}
In this variant, all intermediate rewards are set to 0, effectively removing the step-grained reward signal. As shown in~\autoref{fig:ablation}, this leads to a clear performance drop, confirming the importance of providing informative feedback at each intermediate step.

\textbf{w/o Step-grained Optimization.}
Here, we retain step-grained rewards but discard our optimization strategy by training the model using standard PPO on sub-trajectories ending at intermediate steps. As shown in~\autoref{fig:ablation}, this also results in noticeable performance degradation, indicating that traditional PPO fails to fully leverage step-wise dependencies.

Together, these results demonstrate that both step-grained rewards and step-grained optimization are essential for effectively guiding LLMs in multi-step tool use tasks.

\begin{figure*}[t]
    \centering
    \includegraphics[width=0.9\linewidth]{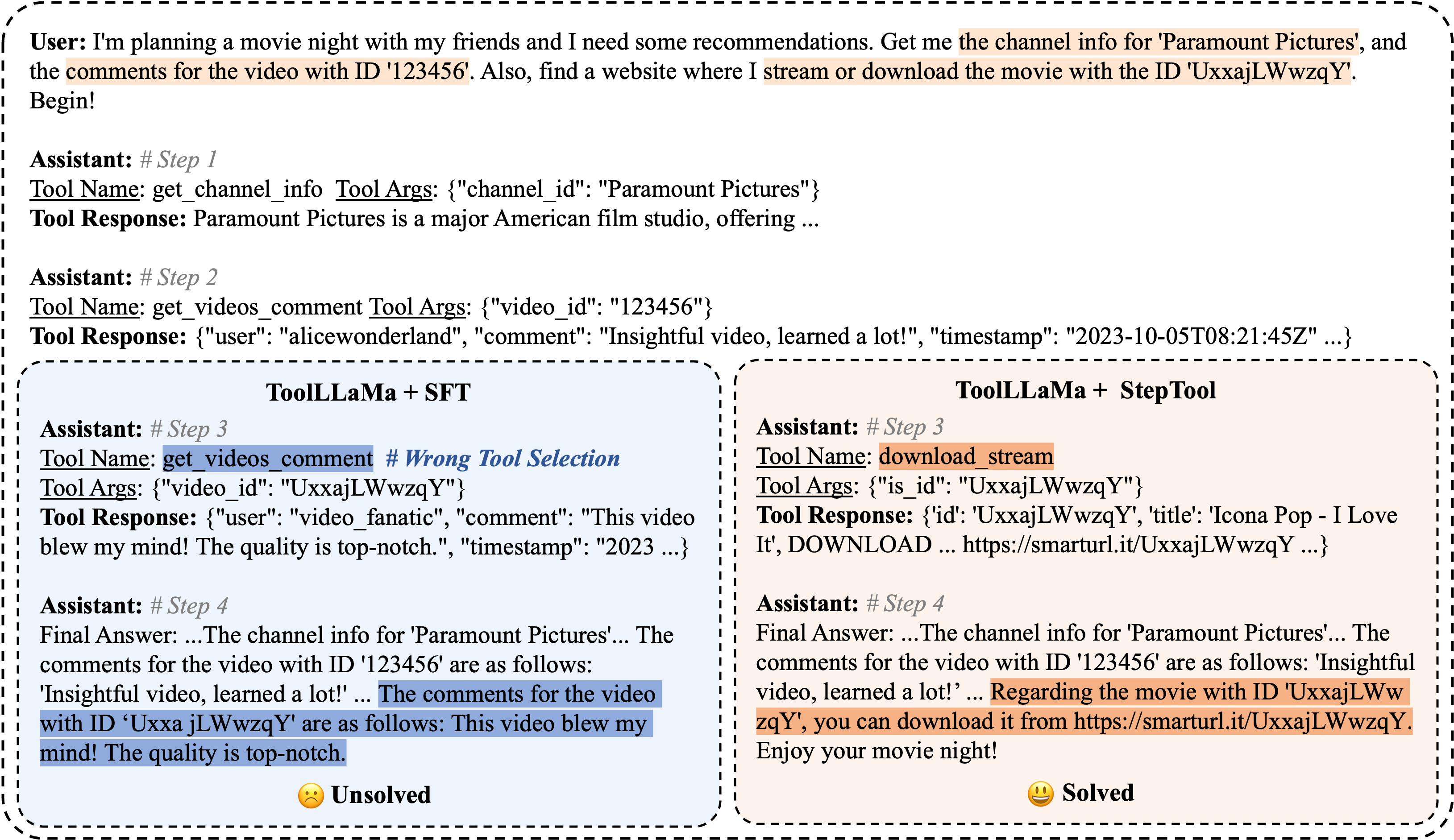}
    \caption{A case study on the I3 Ins. subset comparing ToolLLaMa with SFT and ToolLLaMa with \ourmethod{}: StepTool successfully corrects the wrong tool selection.}
    \label{fig:case_study-0}
\end{figure*}

\subsection{Further Analysis}

\subsubsection{\textbf{Error Analysis}}

\begin{table}[t]
\centering
\caption{The statistics of failure cases for RFT and \ourmethod{} on ToolBench, categorized by error type.}
\begin{tabular}{>{\raggedright}m{5.6cm} m{0.8cm} m{1.0cm}<{\centering}} 
\toprule
\textbf{Error type} &  \textbf{RFT} &  \textbf{\ourmethod{}} \\
\midrule 
\textit{\textbf{\# 1. Tool Selection error}}: The model selects incorrect tools or fails to invoke necessary tools for completing the task. & 12.01\% & \textbf{7.86\%} \\
\textit{\textbf{\# 2. Hallucination Tool error}}: The model fabricates tools that do not exist or are not part of the available toolset. & 2.15\% & \textbf{1.00\%} \\
\textit{\textbf{\# 3. Arguments error}}: The model invokes the correct tool but provides incorrect or irrelevant arguments, leading to failure. & 1.60\% & \textbf{0.96\%}  \\
\textit{\textbf{\# 4. Parse error}}: The tool call is malformed or syntactically incorrect, preventing proper execution. & 1.12\% & \textbf{0.24\%} \\

\bottomrule
\end{tabular}
\label{tab:errors}
\end{table}
To understand model failures in multi-step tool usage, we categorize the error cases for RFT and \ourmethod{} on ToolBench into four types: Tool Selection, Hallucination Tool, Arguments, and Parse errors, as shown in~\autoref{tab:errors}.

Compared to RFT, \ourmethod{} achieves substantial reductions in all error categories. Notably, the \textit{Tool Selection} error rate drops from 12.01\% to 7.86\%, indicating that step-grained optimization significantly improves tool choices. \textit{Hallucination Tool} and \textit{Parse} errors also decline, confirming our method's effectiveness in tool call reliability and precision.

Despite improvements, \textit{Tool Selection} remains the most frequent error for both models, highlighting tool choice as the core challenge in tool-augmented decision-making.

\subsubsection{\textbf{Case Study.}}

To better understand how \ourmethod{} enhances intermediate decisions, we present a representative case in~\autoref{fig:case_study-0}, where it successfully corrects a tool selection error made by ToolLLaMa. 
In this example, the user requests channel information, video comments, and streaming sources for movies. ToolLLaMa first retrieves the correct channel info and video comments, but mistakenly calls the \sethlcolor{wrong_action!90}\hl{get\_videos\_comment} tool again instead of switching to the \sethlcolor{right_action!90}\hl{download\_stream} tool.
After applying \ourmethod{}, the model correctly uses the \sethlcolor{right_action!90}\hl{download\_stream} tool, providing the streaming link and fulfilling the request.
Additional cases of missing tool calls and argument errors are provided in our code repository\footnote{https://github.com/yuyq18/StepTool}.

\section{Conclusion}

In this work, we introduced \textbf{\ourmethod{}}, a step-grained reinforcement learning framework designed to enhance LLMs' ability to tackle complex, multi-step tasks using multiple tools. 
\ourmethod{} features two core components: \textit{Step-grained Reward Shaping}, which evaluates tool invocation success and task contributions, and \textit{Step-grained Optimization}, a step-grained reinforcement-based optimization method. 
Experimental results demonstrate that \ourmethod{} consistently outperforms strong SFT- and RL-based baselines across multiple benchmarks, models, and reasoning strategies, with clear improvements in both task success (Pass Rate) and tool-use accuracy (Recall).
By explicitly optimizing intermediate decisions, \ourmethod{} offers a new perspective for aligning tool-augmented LLMs with step-wise objectives and can be extended to adaptive reward modeling and broader tool-use contexts.

\section{GenAI Usage Disclosure}

During the preparation of this paper, the authors used OpenAI's ChatGPT (GPT-4) to assist with grammar and phrasing. The tool was used similarly to conventional writing aids and did not contribute to the core methodology, experiments, or results.
Although our framework supports flexible reward annotation strategies, we employed GPT-4 to assist with step-level reward annotation, leveraging its strong semantic understanding, accessibility, and cost-effectiveness. In line with ToolBench~\cite{qin2023toolllm}, GPT-4 was also used to evaluate pass rate and win rate following standard protocols. Further details are provided in~\autoref{sec: reward_shaping} and ~\autoref{sec: experiment}.

\bibliographystyle{ACM-Reference-Format}
\bibliography{sample-base}


\end{document}